\documentclass[runningheads]{llncs}
\usepackage[T1]{fontenc}
\usepackage{graphicx}
\usepackage{amsmath,amssymb}

\begin{document}

\title{COAST: Context-Aware Differential Learning for Gene Expression Prediction in Spatial Transcriptomics}
\titlerunning{COAST}

\author{
Keunho Byeon\inst{1} \and
Sunhong Park\inst{1} \and
Jeewoo Lim\inst{1} \and
Jin Tae Kwak\inst{1}
}

\authorrunning{K. Byeon et al.}

\institute{School of Electrical Engineering, Korea University, Seoul 02841, Republic of Korea \\ 
\email{\{bkh5922,sunhongpapa,jeewoolim,jkwak\}@korea.ac.kr}}

\maketitle

\begin{abstract}
Spatial transcriptomics enables profiling of spatial gene expression but is limited by high cost and low throughput, motivating prediction from H\&E histopathology images. Existing context-aware methods mainly supervise absolute expression, while relative expression relationships between spots are rarely used explicitly.
We propose COAST, a context-aware differential learning framework for spatial gene expression prediction. 
COAST conditions the local and global context features with type-specific modulation and aggregates the target and context spot tokens using a Transformer encoder to capture both fine-grained local patterns and slide-level structure.
It is trained with a joint objective that combines absolute expression regression with signed differential regression between the target and context spots.
Experiments on multiple spatial transcriptomics datasets show consistent improvements in correlation- and distribution-based metrics, demonstrating the effectiveness of context-aware differential learning for histology-based spatial gene expression prediction.
\keywords{Spatial Transcriptomics \and Differential Learning \and Context-Aware}
\end{abstract}

\section{Introduction}
Spatial transcriptomics (ST) has emerged as a transformative technology that bridges morphological observations with molecular profiling by providing gene expression measurements within their native spatial context. Despite its great potential, current ST platforms, such as 10x Genomics Visium and Slide-seq, are constrained by high experimental costs, limited resolution, and relatively low throughput. These practical barriers severely restrict their widespread adoption in routine clinical workflows. Consequently, there is growing interest in developing and deploying computational approaches capable of inferring spatial gene expression directly from universally available and cost-effective hematoxylin and eosin (H\&E) stained histopathology images.
Inferring gene expression from histology is inherently challenging due to the complex and highly non-linear relationship between tissue morphology and transcriptomic states. Early approaches framed this as an independent patch-level regression problem, focusing on local feature representations extracted from individual image patches and regressing expression values at each spot in isolation. 
However, gene expression is fundamentally shaped by spatial context. Morphologically similar regions can exhibit distinct expression levels depending on their specific microenvironment, structural compartment, and global tissue composition. To address this, recent methods introduced neighbor aggregation, graph-based propagation, or attention-based mechanisms to incorporate spatial context from adjacent regions.

Despite these advancements, two critical limitations persist. First, most existing models are optimized to regress the absolute expression magnitude at each spot, making predictions sensitive to slide-level variability and protocol-dependent shifts. 
Second, although spatial context is often provided as input, models are rarely supervised to explicitly preserve relative spatial relationships. In biological tissues, spatial gradients and relative changes between regions (such as tumor-stroma boundaries) frequently encode more stable and structurally meaningful signals than absolute counts.

To address these limitations, we propose \textbf{COAST}, a context-aware differential learning framework for spatial gene expression prediction from H\&E histopathology images. 
COAST seamlessly integrates heterogeneous contextual information through type-specific contextual feature modulation, enabling adaptive conditioning of local and global context features. 
Critically, COAST is optimized under a joint objective that combines standard absolute expression regression with signed differential prediction between a target spot and its contextual spots, thereby directly supervising spatial relationships. 
Experiments across multiple public ST datasets demonstrate consistent improvements over existing baselines. Furthermore, the reconstructed gene expression representations retain clinically meaningful prognostic information in downstream survival prediction, highlighting the practical relevance of context-aware differential learning.

\section{Related Works}

\subsection{Histology-based Gene Expression Prediction}
Several studies have investigated predicting spatial gene expression directly from H\&E histopathology images.
Early approaches like STNet~\cite{he2020integrating} introduced convolutional encoders to predict spot-level gene expression from local image patches.
Subsequent methods sought to capture broader tissue architecture by integrating spatial positional encoding and Transformers (Hist2ST~\cite{zeng2022spatial}), constructing spatial graphs over spots for neighborhood-aware prediction (TCGN~\cite{xiao2024transformer}), and employing retrieval-based transfer modeling with graph neural networks (EGNv2~\cite{yang2024spatial}). More recently, NH2ST~\cite{qu2025spatially} introduced a dual-branch architecture with cross-attention and contrastive learning, while PEKA~\cite{pan2025teaching} leveraged a single-cell foundation model for parameter-efficient knowledge transfer. 
Although these methods progressively incorporated spatial context, their learning objectives primarily remain focused on absolute spot-wise expression regression.

\subsection{Relative and Differential Learning Objectives}
Relative and differential supervision has become a cornerstone of robust representation learning.
Methods such as Lifted Structured Embedding~\cite{oh2016deep} and Ranked List Loss~\cite{wang2019ranked} enforce relational constraints among samples.
DIOR-ViT~\cite{lee2025dior} introduced a differential ordinal regression objective that explicitly models ordered relationships between samples to improve structural consistency and robustness.
However, explicitly supervising signed expression differences between spatially separated regions remains largely unexplored in histology-based ST.

\subsection{Feature-wise Affine Modulation}
Feature-wise affine modulation has been widely adopted for conditioning intermediate representations. 
Representative methods include AdaIN~\cite{huang2017arbitrary}, FiLM~\cite{perez2018film}, and SPADE~\cite{park2019semantic}, which apply learned scaling and shifting parameters to adapt features based on external signals.

\begin{figure}[t]
\centering
\includegraphics[width=\textwidth]{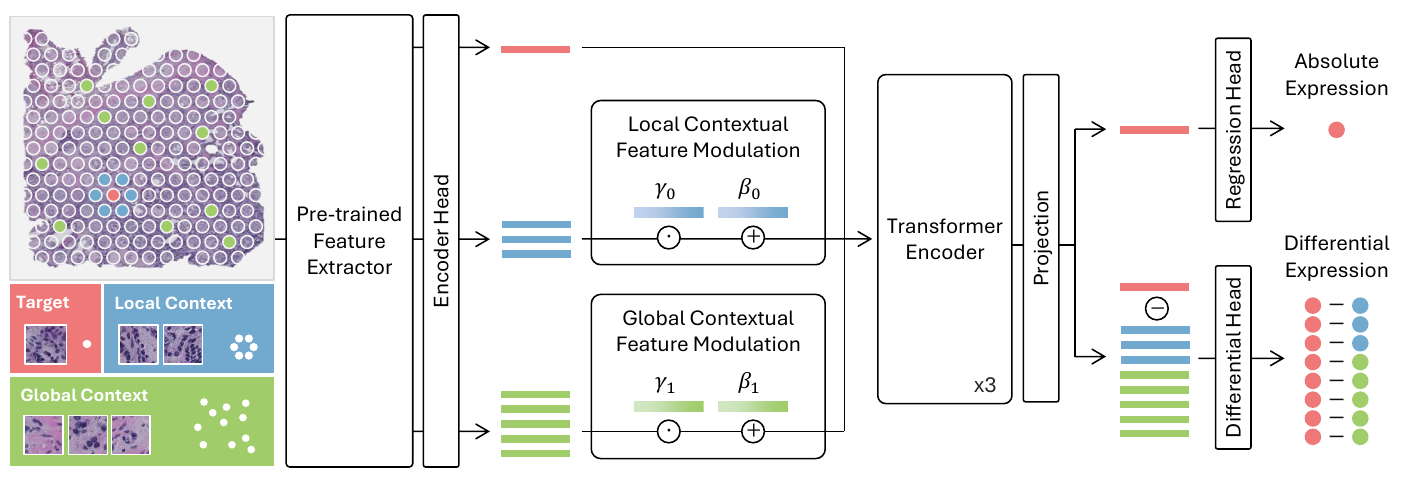}
\caption{
Overview of the COAST framework. COAST integrates local and global context spots through contextual feature modulation and Transformer-based aggregation. The architecture jointly optimizes absolute and differential regression.
} \label{fig:model}
\end{figure}
\section{Methods}

The COAST framework is designed to learn the gene expression patterns at a target spot through explicit interactions with its heterogeneous spatial context (Fig.~\ref{fig:model}). The framework consists of four major components: (1) a feature extractor, (2) context-specific feature modulation, (3) spatio-relational Transformer, and (4) absolute and differential prediction heads. 

\subsection{Feature Extraction and Context Sampling}
Let a whole slide image (WSI) be defined as $I=\{p_i\}_{i=1}^{N}$, where $p_i$ is the image patch associated with spot $i$ and $N$ is the total number of spots. For a given target spot $t$, let $x_t \in \mathbb{R}^D$ denote its feature vector, extracted using a pre-trained feature extractor. 
To effectively capture both the tissue microenvironment and global structure, we sample a context feature set $\mathcal{C}_t = \{x_1, \dots, x_K\}$ containing $K$ spot features from $I$. 
This context set consists of two distinct subsets: $\mathcal{C}_{t} = \mathcal{C}_{\mathrm{L}} \cup \mathcal{C}_{\mathrm{G}}$, where $\mathcal{C}_{\mathrm{L}}$ contains $K_{\mathrm{L}}$ local context spot features sampled via $k$-nearest neighbors, $\mathcal{C}_{\mathrm{G}}$ comprises $K_{\mathrm{G}}$ global context spot features, and $K=K_{\mathrm{L}}+K_{\mathrm{G}}$. 
To construct $\mathcal{C}_{\mathrm{G}}$ during training, we randomly sample spot features from $I$ to enhance robustness and diversity. During inference, we employ a deterministic stratified grid strategy to ensure reproducible and representative context selection.
Finally, each context spot $x_k \in \mathcal{C}_t$ is associated with a type indicator $c_k \in \{0, 1\}$, where 0 and 1 represent local and global types, respectively.

\subsection{Context-Specific Modulation and Spatio-Relational Transformer}
Both target and context spot features are projected into a latent space using a shared encoder head $\Phi_{head}$, which consists of a linear layer followed by a LeakyReLU activation and dropout:
\begin{equation}
    h_t = \Phi_{head}(x_t), \quad h_k = \Phi_{head}(x_k)
\end{equation}
where $h_t, h_k \in \mathbb{R}^d$. 
Although context spot features are sampled from spatially proximal (local) and distant (global) regions, the individual features $x_k$ only carry localized visual information, without retaining their spatial context. Accordingly, naively aggregating local and global context with the target spot can conflate heterogeneous signals and weaken type-specific contributions during context integration.
To effectively integrate these heterogeneous context sources and resolve this spatial ambiguity, we employ a contextual feature modulation mechanism with two sets of scale ($\gamma$) and shift ($\beta$) parameters: $\gamma_{0},\beta_{0} \in \mathbb{R}^d$ for local context ($c_k=0$) and $\gamma_{1},\beta_{1} \in \mathbb{R}^d$ for global context ($c_k=1$). We learn these context-specific scale and shift parameters via embedding layers to modulate the context features:
\begin{equation}
    \tilde{h}_k = h_k \odot (1 + \gamma_{c_k}) + \beta_{c_k}
\end{equation}
where $\odot$ denotes element-wise multiplication. This allows the model to adaptively condition the latent representations based on their spatial context type.

We then form a token sequence by concatenating the target token and the modulated context tokens, $[h_t; \tilde{h}_1,\dots,\tilde{h}_K]$, and feed it into a standard $L$-layer Transformer encoder to aggregate heterogeneous context. The Transformer encoder outputs refined context-aware representations $Z = [z_t; z_1,\dots,z_K]$.

\subsection{Absolute and Differential Prediction Heads}
We project the refined features $Z$ using a shared projection head $\Phi_{proj}$ to obtain the final spot representation $f_t = \Phi_{proj}(z_t) \in \mathbb{R}^D$ and context representations $f_k = \Phi_{proj}(z_k) \in \mathbb{R}^D$. The projection head $\Phi_{proj}$ includes a sequence of LN, a linear layer, LeakyReLU, and dropout. To capture both absolute and relative expression patterns, we introduce two independent linear prediction heads: (1) Regression Head ($\phi_{reg}$): Predicts the absolute gene expression vector $\hat{y}_t = \phi_{reg}(f_t)$ and (2) Differential Head ($\phi_{diff}$): Predicts the expression difference between the target and each context spot from their feature difference, $\Delta \hat{y}_{t,k} = \phi_{diff}(f_t - f_k)$, for $k=1,\cdots,K$.

\subsection{Training Objectives}
The total objective function $\mathcal{L}$ is defined as the weighted sum of the absolute regression loss $\mathcal{L}_{reg}$ and the differential regression loss $\mathcal{L}_{diff}$:
\begin{equation}
    \mathcal{L} = \underbrace{\| y_t - \hat{y}_t \|^2}_{\mathcal{L}_{reg}} + w_{\mathrm{diff}} \frac{1}{K} \sum_{k=1}^{K} \underbrace{\| \Delta y_{t,k} - \Delta \hat{y}_{t,k} \|^2}_{\mathcal{L}_{diff}}
\end{equation}
where $\Delta y_{t,k} = y_t - y_k$ and $y_t$ and $y_k$ are ground-truth expression vectors for the target and context spots, and $w_{\mathrm{diff}}$ is a hyperparameter. This explicit supervision of relative differences encourages the model to learn stable spatial gradients of gene expression, enhancing robustness against slide-level batch effects.

\section{Experiments}

\subsection{Implementation Details}
We employed UNI~\cite{chen2024towards} as a frozen pre-trained feature extractor.
COAST used an input hidden dimension $D{=}1024$, Transformer $L{=}3$ layers, embedding dimension $d{=}512$ and $H{=}8$ attention heads.
We set $K_{\mathrm{L}}=18$ and $K_{\mathrm{G}}=72$ ($K=90$ total) and set the differential weight to $w_{\mathrm{diff}}=1.0$.
The model was optimized with Adam (learning rate $2\times10^{-4}$, weight decay $1\times10^{-4}$) for 10 epochs with batch size 1, selecting the best checkpoint on validation PCC.
All experiments were conducted on a single NVIDIA H200 GPU using Python 3.12 and PyTorch 2.7.

\begin{table}[t]
\caption{
    Comparison of gene expression prediction performance across different models.
    Best results are shown in \textbf{bold}.
}\label{tab:results}
\footnotesize
\resizebox{\textwidth}{!}{
\begin{tabular}{l|ccccc}
\hline
Model 
& PCC $\uparrow$ & MI $\uparrow$ & AUC$_{0\mathrm{vNZ}}$ $\uparrow$ & AUC$_{Q50}$ $\uparrow$ & NRMSE $\downarrow$ \\
\hline
STNet   & $0.0091 \pm 0.01$ & $0.0192 \pm 0.01$ & $0.4784 \pm 0.04$ & $0.5024 \pm 0.01$ & $0.1829 \pm 0.04$ \\
Hist2ST & $0.0516 \pm 0.00$ & $0.0788 \pm 0.03$ & $0.4933 \pm 0.02$ & $0.5000 \pm 0.00$ & $0.1528 \pm 0.03$ \\
TCGN    & $0.1920 \pm 0.09$ & $0.0628 \pm 0.02$ & $0.5571 \pm 0.07$ & $0.6066 \pm 0.05$ & $0.1961 \pm 0.06$ \\
EGNv2   & $0.2469 \pm 0.08$ & $0.0739 \pm 0.03$ & $0.6132 \pm 0.04$ & $0.6453 \pm 0.05$ & $0.1480 \pm 0.04$ \\
NH2ST   & $0.3279 \pm 0.07$ & $0.1011 \pm 0.02$ & $0.6149 \pm 0.07$ & $0.6633 \pm 0.03$ & $0.1449 \pm 0.04$ \\
PEKA    & $0.3863 \pm 0.08$ & $0.1109 \pm 0.03$ & $0.6508 \pm 0.04$ & $0.6879 \pm 0.03$ & $0.1322 \pm 0.02$ \\
\hline
\textbf{COAST} & $\mathbf{0.4262 \pm 0.08}$ & $\mathbf{0.1487 \pm 0.03}$ & $\mathbf{0.6567 \pm 0.06}$ & $\mathbf{0.7170 \pm 0.02}$ & $\mathbf{0.1245 \pm 0.02}$ \\
\hline
\end{tabular}
}
\end{table}

\subsection{Datasets}
SpaRED~\cite{mejia2024enhancing} provides multiple spatial transcriptomics cohorts collected from diverse tissues and species.
In this study, we used seven SpaRED datasets that include predefined test splits and contain 128 target genes: Abalo Human Squamous Cell Carcinoma (AHSCC)~\cite{abalo2021human}, Erickson Human Prostate Cancer P1 (EHPCP1)~\cite{erickson2022spatially}, Mirzazadeh Mouse Bone (MMBO)~\cite{mirzazadeh2023spatially}, Mirzazadeh Mouse Brain P1 (MMBP1) and P2 (MMBP2)~\cite{mirzazadeh2023spatially}, Vicari Mouse Brain (VMB)~\cite{vicari2024spatial}, and Villacampa Lung Organoid (VLO)~\cite{villacampa2021genome}. 
For each spot, we directly used the provided $224{\times}224$ H\&E patch and the Transcripts Per Million (TPM) normalized, log1p-transformed expression values released in SpaRED.

\subsection{Comparison Results}
\begin{figure}[t]
\centering
\includegraphics[width=0.95\textwidth]{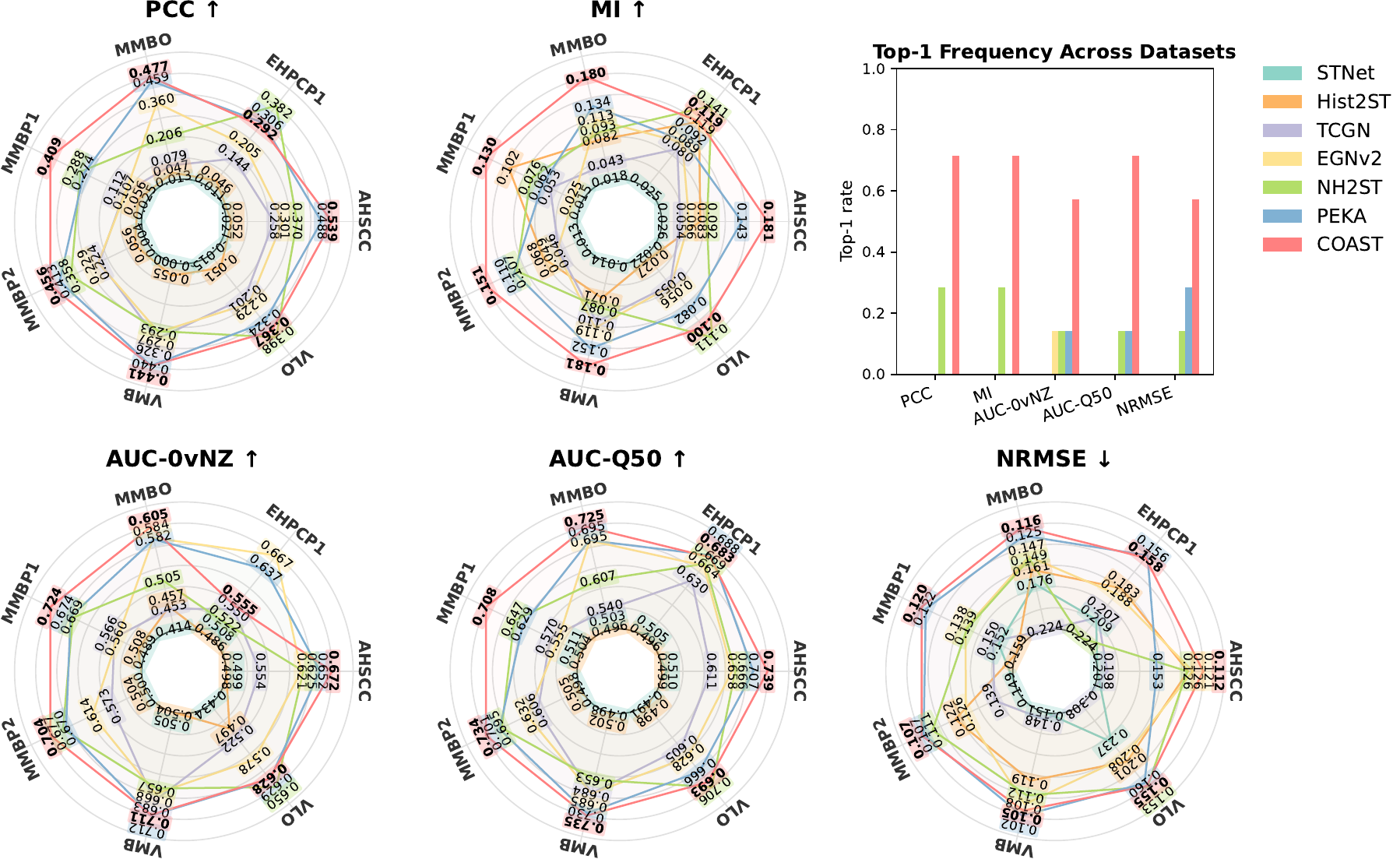}
\caption{
Comparison of per-dataset gene expression prediction performance using radar charts and Top-1 frequency rates. 
} 
\label{fig:comparison_all_results}
\end{figure}

\begin{figure}[t]
\centering
\includegraphics[width=0.92\textwidth]{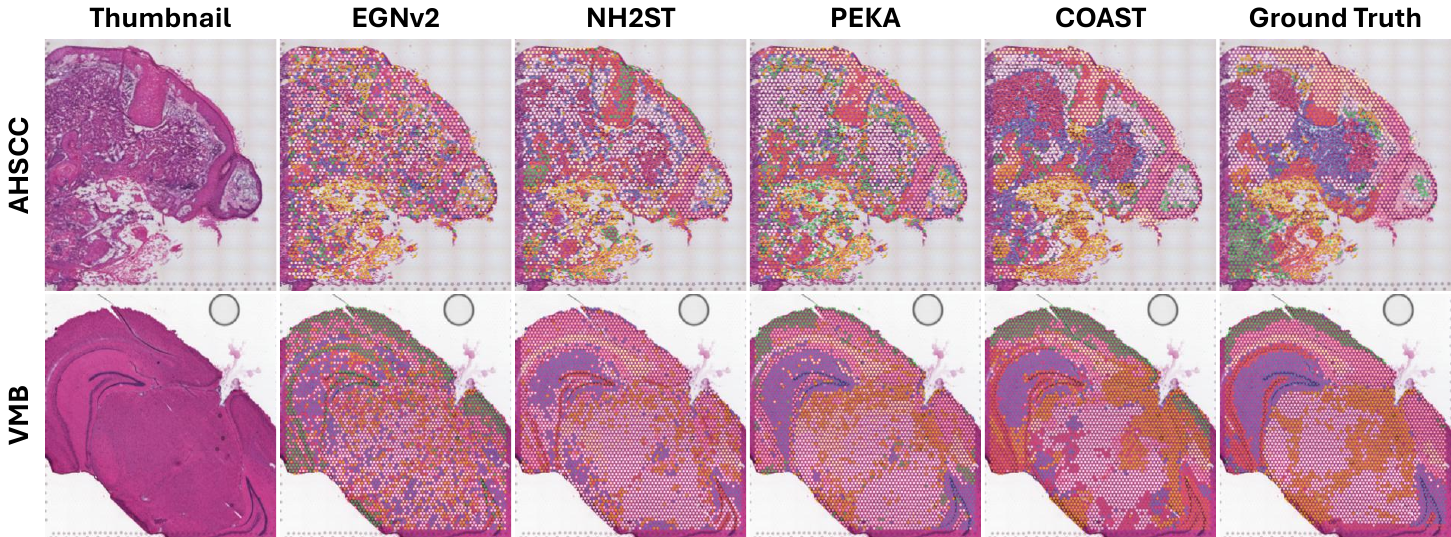}
\caption{
Visualization of Leiden clustering on predicted and ground-truth expression.
}
\label{fig:comparison_results_leiden}
\end{figure}

We compared COAST with six representative and competitive baseline models, including STNet, Hist2ST, TCGN, EGNv2, NH2ST, and PEKA. 
For baseline models, we used the official implementations and followed their default hyperparameter settings.
Table~\ref{tab:results} summarizes the average performance across seven SpaRED datasets.
We report gene-wise averages over $G=128$ genes using Pearson Correlation Coefficient (PCC), Mutual Information (MI) estimated with k-nearest neighbors ($k=5$), ROC-AUC$_{0\mathrm{vNZ}}$ for distinguishing zero from non-zero expression, ROC-AUC$_{Q50}$ for discriminating values above and below the median, and Normalized Root Mean Square Error (NRMSE).
COAST achieved the best performance on all five evaluation metrics, obtaining substantial performance gains over all baselines: 0.0399 $\sim$ 0.4171 in PCC, 0.0378 $\sim$ 0.1295 in MI, 0.0059 $\sim$ 0.1783 in AUC$_{0\mathrm{vNZ}}$, 0.0291 $\sim$ 0.2170 in AUC$_{Q50}$, and 0.0077 $\sim$ 0.0716 in NRMSE.
We further analyzed per-dataset performance using radar charts and Top-1 frequency rates, measuring how often a method achieves the best result among all competing models (Fig.~\ref{fig:comparison_all_results}). COAST consistently formed the largest polygons in radar charts and obtained the highest Top-1 frequency, indicating superior and balanced performance across diverse datasets and evaluation metrics. Qualitative assessment via Leiden clustering on reconstructed expression (Fig.~\ref{fig:comparison_results_leiden}) further reveals that COAST captures more coherent spatial domains, reflecting improved structural consistency.

\subsection{Ablation Study}
Table~\ref{tab:ablation_all} demonstrates ablation experiments using PCC to validate three key components of COAST: (1) context sampling size ($K_{\mathrm{L}}$ and $K_{\mathrm{G}}$), (2) a differential objective weight $w_{\mathrm{diff}}$, and (3) context-specific modulation.
For context sampling, the combination $K_{\mathrm{L}}{=}18$ and $K_{\mathrm{G}}{=}72$ yielded the best performance (PCC=$0.4262$), whereas increasing either $K_{\mathrm{L}}$ or $K_{\mathrm{G}}$ degraded performance. 
The ablation of $w_{\mathrm{diff}}$ revealed the critical role of differential learning. Removing the differential objective ($w_{\mathrm{diff}}=0$) led to a substantial decrease in PCC by $0.0477$. While optimal performance was achieved at $w_{\mathrm{diff}}{=}1.0$, increasing the weight to 2.0 resulted in a performance drop.
Regarding context-specific modulation, jointly applying scaling and shifting ($\gamma \text{ and } \beta$) achieved the strongest performance.
Overall, these results suggest that COAST’s gains are not driven by a single component but arise from complementary design choices. Across these three experiments, most ablation variants remained superior to PEKA (the strongest baseline), further indicating the superiority of COAST.

\begin{table}[t]
\centering
\caption{Ablation study on context sampling, differential objective weight, and contextual feature modulation strategy.}
\label{tab:ablation_all}
\footnotesize
\begin{tabular}{c c|c||c|c||c|c}
\hline
$K_{\mathrm{L}}$ & $K_{\mathrm{G}}$ & PCC $\uparrow$ &
$w_{\mathrm{diff}}$ & PCC $\uparrow$ &
Modulation & PCC $\uparrow$ \\
\hline
18 & 36 & $0.4178 \pm 0.07$ &
0.0 & $0.3785 \pm 0.09$ &
w/o & $0.3999 \pm 0.08$ \\

18 & 108 & $0.4214 \pm 0.07$ &
0.5 & $0.3762 \pm 0.09$ &
$\gamma$ & $0.4128 \pm 0.07$ \\

36 & 72 & $0.3926 \pm 0.13$ &
2.0 & $0.3914 \pm 0.12$ &
$\beta$ & $0.4131 \pm 0.07$ \\
\hline
18 & 72 & $0.4262 \pm 0.08$ &
1.0 & $0.4262 \pm 0.08$ &
$\gamma \text{ and } \beta$ & $0.4262 \pm 0.08$ \\
\hline
\end{tabular}
\end{table}

\subsection{Downstream Task: Overall Survival Prediction on TCGA-LUAD}
To evaluate whether the gene expression representations reconstructed by COAST preserve clinically relevant prognostic signals, we conducted overall survival (OS) prediction on the TCGA-LUAD cohort.
Patient-level data splits were used with 5-fold cross-validation. 
Patch-level image features were extracted using the frozen UNI~\cite{chen2024towards}, and gene expression representations were inferred from a model pretrained on the VLO dataset without fine-tuning. We employed CLAM-SB~\cite{lu2021data} for slide-level risk prediction and optimized the model using the Cox proportional hazards partial likelihood~\cite{cox1972regression}.

We compared three input configurations: (1) image features only (Img-Only), (2) image features combined with PEKA-predicted expression representations (Img+PEKA), and (3) image features combined with COAST-predicted expression representations (Img+COAST). 
For a biological reference, a bulk RNA baseline was trained using a standard Cox proportional hazards model directly on patient-level bulk transcriptomic profiles without image features.
OS prediction performance was quantified using the concordance index (C-index). 
As shown in Table~\ref{tab:downstream_results}, Img+COAST achieved an average C-index of $0.6029$, comparable to bulk RNA ($0.6042$) and outperforming Img-Only ($0.5845$) and Img+PEKA ($0.5904$). 
These results demonstrate that COAST-derived representations capture clinically meaningful prognostic signals from morphology alone.

\begin{table}[t]
\centering
\footnotesize
\caption{
Results of overall survival (OS) prediction on TCGA-LUAD.
}
\label{tab:downstream_results}
\begin{tabular}{c|c c|c c c c}
\hline
Fold & \multicolumn{2}{c|}{\#Patients (\#Slides)} & \multicolumn{4}{c}{C-Index $\uparrow$} \\
 & Event & Censored & Bulk RNA & Img-Only & Img+PEKA & Img+COAST \\
\hline
0 & 33 (34) & 59 (70) & 0.5839 & 0.5019 & 0.5034 & 0.5146 \\
1 & 33 (50) & 58 (58) & 0.6005 & 0.7134 & 0.6990 & 0.6986 \\
2 & 32 (34) & 59 (63) & 0.5423 & 0.3949 & 0.4409 & 0.4978 \\
3 & 32 (47) & 59 (63) & 0.6356 & 0.6635 & 0.6580 & 0.6560 \\
4 & 32 (41) & 59 (59) & 0.6588 & 0.6488 & 0.6508 & 0.6473 \\
\hline
Avg. & 162 (206) & 294 (313) & $0.6042\pm0.05$ & $0.5845\pm0.13$ & $0.5904\pm0.11$ & $0.6029\pm0.09$ \\
\hline
\end{tabular}%
\end{table}

\section{Conclusion}
In this paper, we present COAST, a context-aware differential learning framework for predicting spatial gene expression from H\&E-stained histopathology images. 
By incorporating both local microenvironment and global tissue structure as heterogeneous context spots, COAST effectively overcomes the limitations of existing methods that primarily focused on patch-level regression. 
Across seven datasets, COAST consistently outperforms representative baselines. Beyond prediction accuracy, COAST-derived expression representations preserve clinically meaningful prognostic signals, improving OS prediction on TCGA-LUAD when combined with histology features, comparable to actual bulk RNA sequencing data.
These findings suggest that context-aware differential learning is an effective strategy for enhancing histology-based spatial gene expression prediction.

\begin{credits}
\subsubsection{\ackname}
This work was supported by a grant of the National Research Foundation of Korea (NRF) (No. RS-2025-00558322 and RS-2024-00397293).

\end{credits}

\bibliographystyle{splncs04}
\bibliography{main}

\end{document}